\documentclass[conference]{IEEEtran}
\usepackage{cite}

\ifCLASSINFOpdf
   \usepackage[pdftex]{graphicx}
\else
\fi
\usepackage{amsmath}
\usepackage{etoolbox}
\usepackage{adjustbox}
\begin{document}
%
\title{A Study of Time-varying Cost Parameter Estimation Methods in Automated Transportation Systems based on Mobile Robots}

\author{\IEEEauthorblockN{Pragna Das and Llu{\'\i}s Ribas-Xirgo }
\IEEEauthorblockA{Department of Microelectronics and Electronic Systems\\
Universitat Aut{\`o}noma de Barcelona\\
Bellaterra, Barcelona 08193\\
Email: \{Pragna.Das,Lluis.Ribas\}@uab.cat}
}


%


\maketitle

\begin{abstract}
Control of systems of automated guided vehicles involves action planning at many levels. For efficient control of these systems, accurate estimation of cost parameters (speed, energy, task completion performance, \textit{et~cetera} is required. These parameters change along time, particularly in battery-operated robots, which are very sensitive to battery level variations. This work addresses the problem of on-line cost parameter identification and estimation for proper control decisions of the individual mobile robots and for the system as a whole. Several filtering and estimation methods have been investigated with respect to travelling times, which are dramatically affected by battery charges and condition of facility's floors, among  other factors. Results show that these parameters depend on the robot, the route and the moment, so they are linked to a particular robot, a region of the floor and a time period (or to a battery level). Moreover, differences with static, pre-runtime travelling time computations, either heuristically or by characterization of real robots, are large enough to affect to system's performance and overall productivity and efficiency.  
\end{abstract}


%
\IEEEpeerreviewmaketitle

\section{Introduction}
Flexible manufacturing plants, warehouses and logistic hubs depend on efficient, robust, adaptable and evolving internal transportation systems \cite{LluisIsmaelMoreno} built upon mobile robots (MRs) or automatic guided vehicles (AGVs). AGVs are driver-less carriers used on the factory floor as mobile production platforms or for material transportation. AGVs are gaining importance in automation, logistics and transportation systems as they provide better flexibility and adaptability essential for automated systems.
From the genesis of the application of AGV in manufacturing and transportation system, supervisory control and data acquisition (SCADA) software systems were used for controlling the AGVs in multi-robot systems \cite{SCADA}. 

%
 

\subsection{Challenges in SCADA solved by Multi-Agent Systems}
Traditional SCADA approaches cannot keep up with today's systems sizes and non-functional requirements like performance, maintainability, scalability, and dependability \cite{abbas2014future} and the requirement of expandability and re-configurability, effective communication and cooperation and fault tolerance at system level. Also, it limits the scope of expansion and enhances the cost of customisation. The central controlling hierarchical organisation may have a shut down with a single failure point. This makes the traditional approaches more fragile with increase overhead costs. The new software paradigm of multi-agent systems (MAS) using all its attributes has the capability to provide solutions to these challenges and to include all the above mentioned requirements in order to implement efficient intelligent manufacturing systems. In this approach, \textit{agents} can represent any physical manufacturing entity like AGVs or tools or operators or aggregation of all of these. 
\begin{figure}[h]
\centering
\includegraphics[width=3.1in]{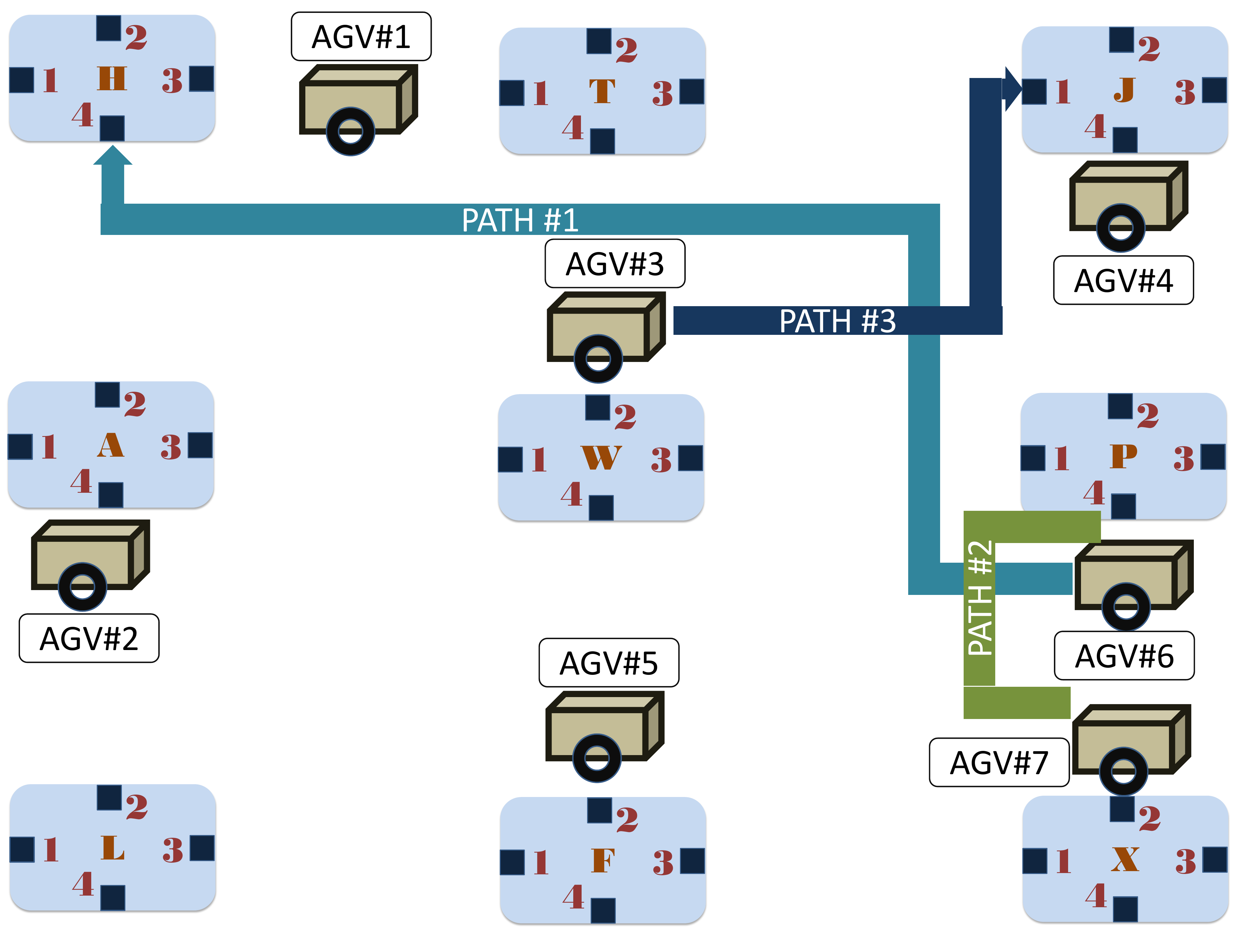}
\caption{An example scenario for the problem}
\label{fig_expl}
\end{figure}
\subsection{Challenges in MAS-based SCADA}


Though MAS-based SCADA reduces or nullify the recurrent customization costs and overhead costs for system failure, this still performs sub-optimally or has limitation due to the following reasons:
\begin{enumerate}
    \item Most systems consider and estimate control parameters heuristically and do not consider updating them in time varying manner
    \item Decisions for one MR (for individual task) or a group of MRs (for co-operative tasks) are taken independent of other MRs operating in same environment 
\end{enumerate}

However in production environment involving multiple MRs, performance of one MR will always depend on other MR's performance. The following example explains such a situation.

Consider Fig \ref{fig_expl}, where an illustration of a system of multiple transportation AGVs is shown. 
Let, for the agent AGV\#6 the path for the movement from port P\#4 to port H\#4 has to be computed. Assume that, at time $t_0$ agent AGV\#6 has its full battery level and can move with its maximum speed. So at that time, the optimal path for AGV\#6 has been computed as PATH\#1. 
But there could be a situation where AGV\#6 has exhausted the battery in its previous task(s) and at time $t_7$ it has to again traverse PATH\#1. Now, it will take a very long time to load at Port P\#4 and traverse that distance from Port P\#4 to Port H\#4 taking that load. 
Moreover due to that AGV\#3 which is supposed to load at Port W\#2 and traverse to Port J\#1 taking the PATH \#3 can now have a collision with AGV \#6 as AGV \#6 is delayed.
Thus at that reduced speed Path\#1 is not the optimal path as the possibility of collision with agent AGV\#3 is there.

However with correct estimation of current speed of AGVs the control decisions in above scenario can be improved as 
1) AGV\#6 with less battery (less speed) can be utilised for some other task now
2) Communicating the status of AGV\#6 to nearby agents e.g AGV\#7 with more battery (higher speed) can be deployed to take Path\#2
to load at Port P\#4 and traverse the distance to Port H\#4 to
carry that load. This will also avoid the possible collision with AGV\#3.




\subsection{Cost Parameters}
\label{cost_para}
The above example shows that to obtain better path planning, parameters considered in existing control architecture is not sufficient. In addition to that controller needs to consider the current speed of individual MR as a control parameter in order to correctly take control decisions. These type of parameters are not limited to only path planning but in other different tasks as well. Similar situations can arise 
in different other tasks e.g uploading a box or any material, capacity of carrying a particular load or accuracy of fitting a mechanical part.

As mentioned earlier, current research works on multi-robot control systems either estimate these parameters at agent level or consider some heuristic cost for system level decision making. However, for efficient control decision making, these parameters needs to be estimated at system level. Thus in this work we are not interested to find optimal path of AGVs or any other particular task rather to study and estimate these type of dynamic coefficients or parameters in general for making better control decisions. As these parameter are time varying in nature and directly influence the cost of the current task (consequently the whole system), these can be modelled as cost parameters to the system. As for the above example we model the current speed by traversal time.

Although, these dynamic coefficients are observable for each AGV at the local robot level, identification and estimation of these parameters at the higher levels helps in decision making to ensure more cost-effective decisions for each AGV and also the system as a whole (for example in the above example, AGV\#6 needs to be replaced by neighbouring AGV\#7 and this decision can not be taken at local robot level). Please note here, that these cost parameters, like any other parameters of AGVs, are also influenced by the battery exhaustion, quality of the shop floor, wear and tear of mechanical parts, conditions of the object to be carried, unknown dynamic obstacle etc and thus tend to have different values at different time instance. Hence, the cost parameters (traversal time for each of the paths in the above example) are continuously evolving.

Also, when the high level decisions like computing optimal path, assigning task or deploying to carry a load are taken, the goal is known very clearly, but there are obscurities for the information on how to achieve that goal as these dynamic coefficients or cost parameters are applied at lower levels of decision taking algorithms (e.g. local controllers). We are trying to bridge this gap where the correct cost parameter estimates evolving over time can be provided for various purposes of automating the processes and reach the correct goal at current time. Therefore, we want to apply them at higher levels.

The experiments shows that these cost parameters vary over several environmental factors as mentioned above. Thus it is easy to follow that correct estimation of these cost parameters can lead to better operational control decisions and path planning of each individual AGV  and the whole system, when compared to a control system based on heuristic cost.

The reported works in the field of multi-robot systems are mostly concerned with the following two categories: 
\begin{itemize}
    \item Problems with the cooperative and collaborative functions like scheduling \cite{nourjou2014dynamic}, task allocation \cite{sung2013improving}, path-planning \cite{ahmed2013multi} 
    \item Problems related to individual robots like localisation \cite{6469211}, dynamic and physical parameters identification and estimating \cite{gautier2013new}, position and orientation estimation \cite{cotugno2013extended} and obstacle estimation \cite{lee2010mobile}
\end{itemize}
 However, none of these proposed any solution for estimation of cost parameters, proposed here.
In this work, unlike the previous work on parameter estimations, we investigate and find an approach to estimate these time-varying cost parameters which are actually derived from the robot level of each AGV as a result of functions done by the actuators and sensors in each AGV.  But we are identifying and estimating them at the higher decision making levels to make better cost efficient functions at the robot level, so as to provide the necessary bridge between these two categories of works. We show that standard estimation methods can be effectively used for correct estimation of these cost parameters.

%


In summary, the contribution of this work can be stated as follows:

1)We propose a different type of parameters (cost parameter) to be used at system level for overall improvement of  performance
2)We find an efficient estimation method among the standard methods for these cost parameters
3)We use a multi-level controller based on MAS derived Agent-Based Modelling (ABM) for this analysis

The rest of the paper is arranged as follows: First we analyse few state of the art research proposals which are similar to our problem in Section \ref{background}, in Section \ref{prototype} we describe a prototype transportation system used as the experimentation platform.
Section \ref{re_prob_def} formulates the problem in the view of our architecture. Some standard methods for estimating parameters are described in section \ref{methods}. A section is devoted to  the experimental verification of our proposal on how to efficiently estimate these cost parameters before concluding with a discussion and further scope.

\section{Background}
\label{background}

Although, there are many recent state of the art works related to multi-robot systems in transportation and automated industry,
dealing with co-operative path-planning, parameter identification of robot dynamics, adaptive controlling through position control, most related to our work are the followings:

In \cite{nestinger2012adaptive}, the authors have proposed a adaptive on-line estimation of system parameters which are time-varying. 
Here, the model parameters of the dynamics of mobile robot of one robot and then of all the N robots of the system are estimated as system parameters to arrive at a coherent estimate. Although the time varying aspect of the model parameters is similar to our work, we are focused on determining to estimate from performance quality or capability of the mobile robots which are impacted by the different environmental factors and which determine the cost of doing task. We call these as cost parameters and these are time-varying in nature. 

The work of  Confessore, Fabiano and Liotta in \cite{Confessore2011} has proposed a minimum cost approach to solve the dispatching problem in an multi-robot system where the network of mobile robots is represented as a graph. The final goal of their work is to determine the minimum cost of performing a task through this graphed network. Like \textit{Confessore, Fabiano and Liotta}, we are also determining the performing ability or quality considered as cost, but our final goal is to use these cost parameters of each AGV in the system to be useful for various decision and controlling purposes. 

Though both the above works can be considered most relevant to the current work, these are not directly comparable as they used different time varying parameters differently. In current work we estimate traversal time of particular AGV as an instance of above mentioned cost parameters. There is no research proposal addressing the same problem as ours. This is the first work to address this type of parameters.

\section{Prototype Platform}
\label{prototype}
 To study and evaluate the estimation methods of cost parameter suggested previously, a scaled prototype platform is required which consists of AGVs in the laboratory set-up for mitigating the gap between the abstract idea and the actual implementation of the system. 
 

We use ABM as the basic control architecture for the individual AGVs for factors like better co-ordination and re-configuration, optimization, consensus and synchronization. In \cite{ribas2013agent}, the authors have proposed ABM for internal transportation systems where agents are categorised into two classes, first for the transportation like AGVs and auxiliary agents such as coordinators and second for rest of the system like physical counterparts (machinery, work-cells, etc.) or software systems (e.g. production management software). 


\subsection{Scaled prototype}

\begin{figure}
\centering
\includegraphics[scale = 0.06]{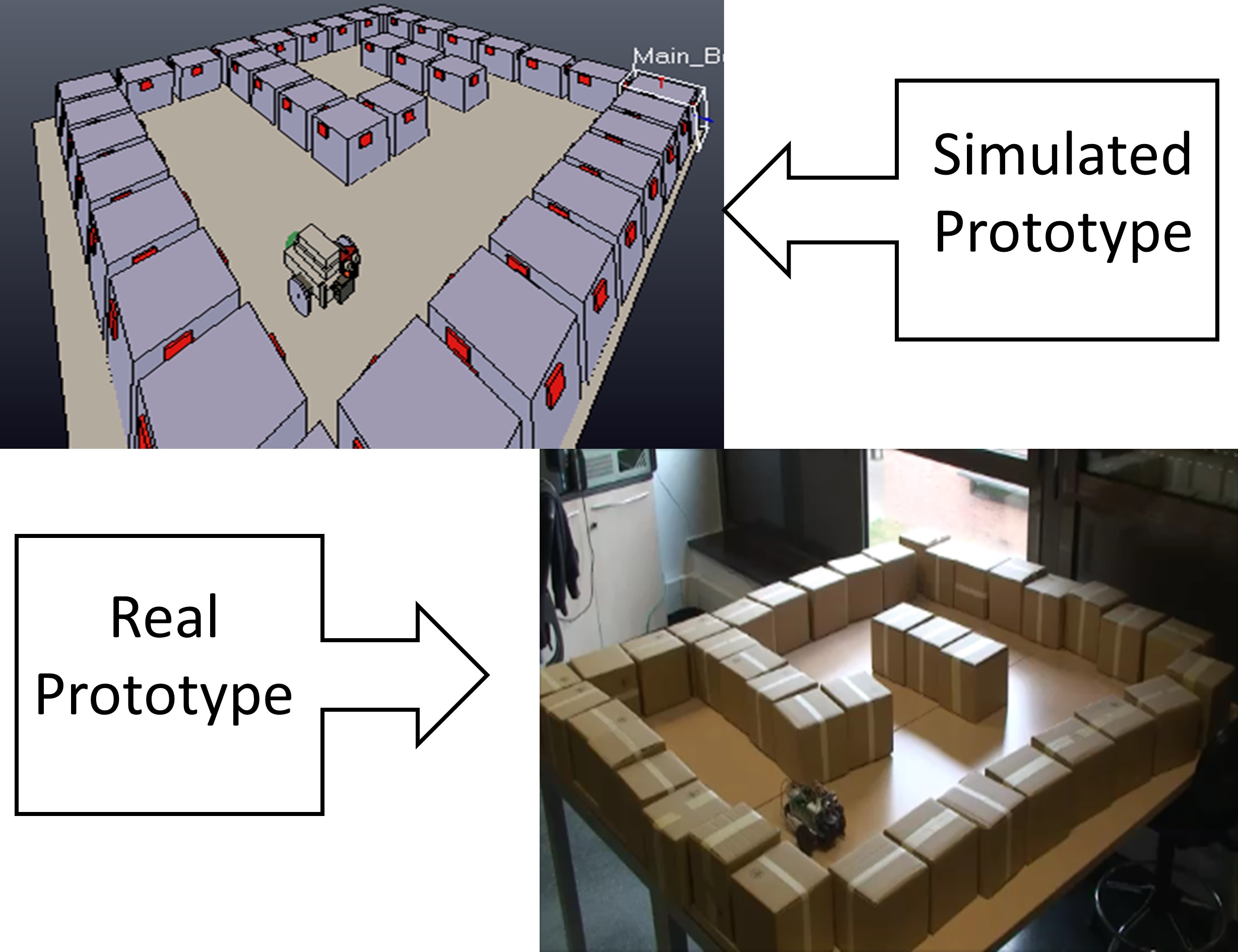}
\caption{The simulated and real scaled prototype}
\label{sim_real}
\end{figure}

The main part of the scaled prototype is a typical replication of a transportation shop floor made of mobile robots and pathways in between with ports to load and unload. The hobby robots, developed using body of the Boebot \cite{parallax} are the transportation agents. The environment is constituted using boxes which build up labyrinth like pathways for the robots to navigate. Also, the whole prototype is simulated in V-Rep software In Figure~\ref{sim_real}, both the simulated and the real prototypes are shown.

\subsection{Controller architecture}
\label{Architecture_Of_Controller}

In this work, we shall focus on the controller architecture of a single AGV. In case of multi-agent systems, the controllers are similar in each of the agents. Our architecture also provides mechanism for communication and sharing of information between the agents\cite{ribas2013agent}. This aspect can be further utilized for updating cost parameters in future.   
\begin{figure}
\centering
  
\includegraphics[scale = 0.4]{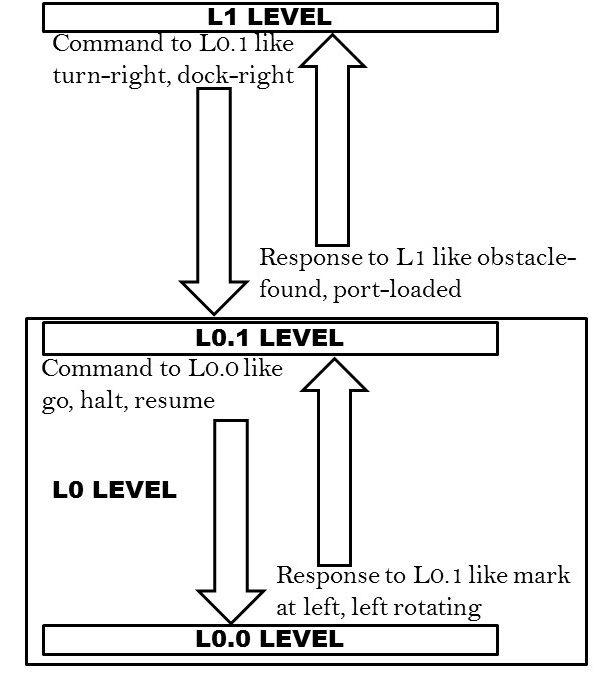}
\caption{Architecture of the controller}
\label{controller}
\end{figure}
The control structure consists of two major control layers (Figure~\ref{controller}), namely $L$1 level and the $L$0 level. The $L$0 level is divided into two sub levels $L$0.1 and $L$0.0 levels respectively. The $L$0 level and $L$1 both functions on each of the agents individually. Here, the $L$0.0 level can only communicate with the $L$0.1 level and can not have any direct communication with the $L$1 level. The $L$0.1 level is the intermediate level which communicates with both $L$0.0 level and $L$1 level. The functions of controlling are more simplified in $L$0.1 and $L$0.0 levels. The $L$1 level is engaged in controlling more complex functions like path planning, task assigning and finding destination poses for each of the robots. In this architecture, the inter-agent communications are done in the highest level ($L$1). Now, each of the robots in our scaled prototype platform have the $L$0.0 (implemented in the micro-controller board of the robot) and $L$0.1 (implemented in the mounted Raspberry-Pi in the robot) levels together implemented in them and each of the $L$1 is to be implemented in a personal computer (PC) with Internet access.
\subsection{Model}
\label{shopfloordesc}
  
\begin{figure}
\centering
\includegraphics[scale = 0.13]{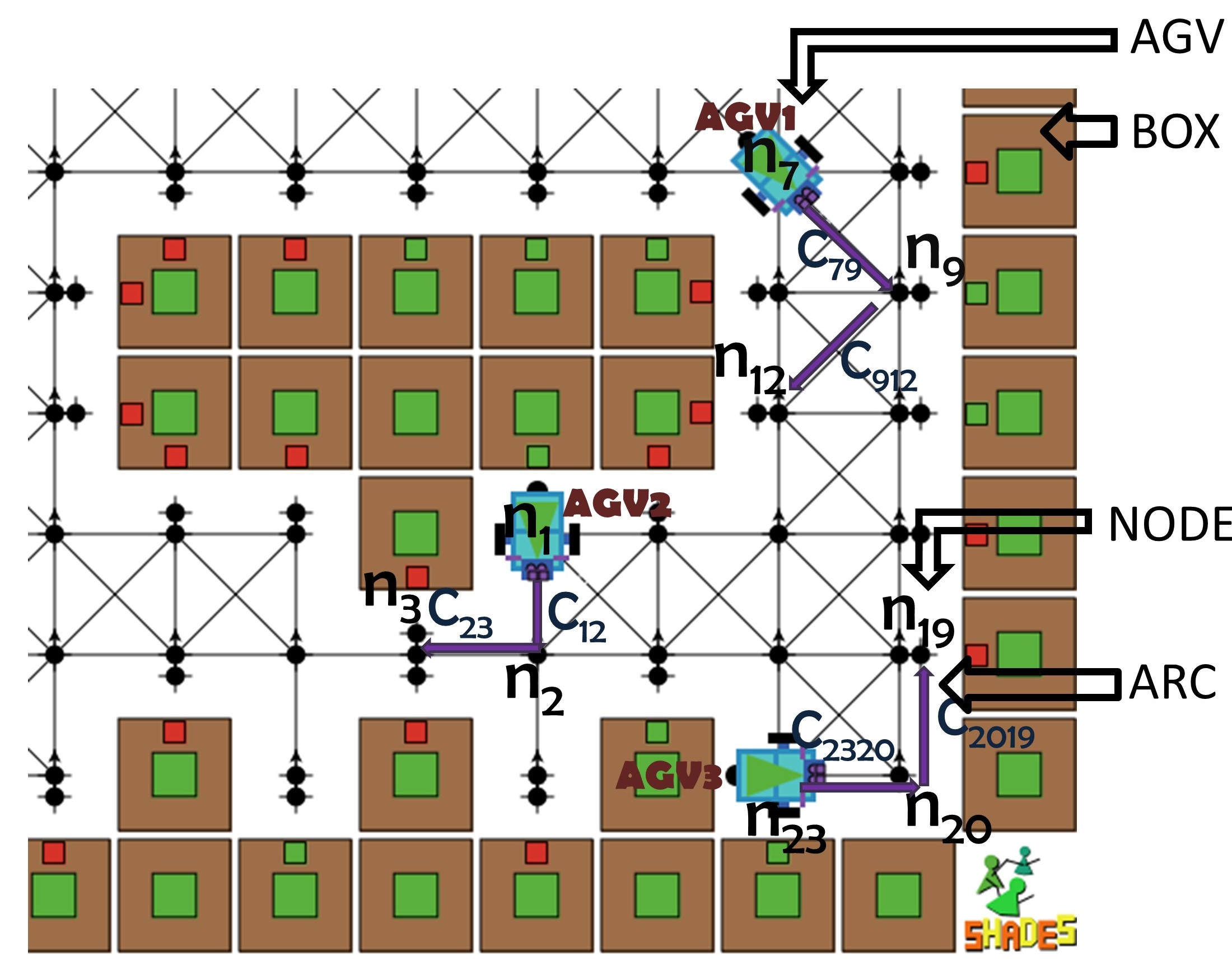}
\caption{The transportation floor}
\label{netlogosample}
\end{figure}
In Figure~\ref{netlogosample}, a simulation of the shop floor has been described. This shop floor is designed as a graph. As denoted in Figure~\ref{netlogosample}, a valid port like $n_3$ is denoted by a node and $n_5$ denotes a bifurcation point where three possible arcs can be taken. The connecting lines ($a_{53}$) between the two nodes $n_5$ and $n_3$ is an arc. 
\section{Formulation of the Problem in light of prototype platform}
\label{re_prob_def}
According to Figure~\ref{netlogosample}, each of the arc in the shop floor is associated with some cost in terms of energy exhaustion, a dynamic obstacle, condition of the floor, load it carries etc as discussed in Section \ref{cost_para}. The time to traverse an arc by a mobile robot (Figure \ref{netlogosample}) is thus conceptualised as cost parameter for that robot. Hence, we formulate that parameter as $C(t,e,b,f,o)$, which is time-varying, where $t$ is for time, $e$ battery state of charge, $b$ is for tire condition, $f$ is for frictional force of the floor, $o$ is for obstacle dependencies.$C(t,e,b,f,o)$ is time-varying from the perspective that at a particular instance of the time, the cost of that particular arc is dependent on battery discharge, condition of the floor or any dynamic obstacle. Hence, for the arc $a_{53}$ in Figure~\ref{netlogosample}, 
the cost parameter is denoted by $C_{53}(t,e,b,f,o)$. Also, for the same robot, for the arc $a_{14}$, the cost parameter will be
$C_{14}(t,e,b,f,o)$. 

Now, a particular path will comprise of one or more arcs. Thus, for a particular path traversal, there are one or more of $C(t,e,b,f,o)$ for a particular robot. Our work focuses on estimating all of such $C(t,e,b,f,o)$ to get the next prediction on the next time instance on-line and recursively update all the prediction values for all the $C(t,e,b,f,o)$ estimates till that time in order to implement a better path planning and control strategy.

\section{Approaches for estimation of parameters}
\label{methods}

 
 We are estimating a time-varying cost parameters which are different from the estimations generally investigated in the state of the art research proposals, in case of robotic systems. 

For parameter identification and estimation, the standard methods include least-square estimator approach \cite{flacco2012line}, least-square moving window method (LSMW) \cite{wang2009fast}, recursive least square (RLS) method \cite{wang2009fast}. Both LSMW and RLS methods are deployed for impedance parameter estimation for robot control in \cite{wang2009fast} .  Further to this, the system state vector was measured at any time and the unknown dynamic parameters are estimated using a Kalman filter (KF) \cite{sujan2003optimal}. 

In LSMW, the data set ($X$, $Y$) of length $L$ is such that,
\begin{equation}\label{eq:Lsmw1}
\ Y = X \theta+W
\end{equation}
where, 
$X^T$=($x_1$,$x_2$,$x_3$,......,$x_L$), 
$Y^T$=($y_1$,$y_2$,$y_3$,......,$y_L$) and $W$ is the measurement noise.

Now, with a window size $l \in N$ such that $l < L$, the number of estimations of $\theta$ will be $L-l$ +1. The estimation is given by,
\begin{equation}\label{eq:Lsmw2}
\ \hat{\theta_i} = X^{\#}_iY_i
\end{equation}
where,
\begin{equation}\label{eq:Lsmw3}
  \ X^{\#}_i = \left( X^T_iX_i \right)^{-1}X^T_i
\end{equation}
and 
$Y^T_i$ = ( $y_i$, $y_{i+1}$, ........., $y_{i+l-1}$),
$X^T_i$ = ( $x_i$, $x_{i+1}$, ........., $x_{i+l-1}$),
$i$ = 1,2,......,$L-l+1$
with the estimation error
\begin{equation}\label{eq:Lsmw4}
  \ \hat{e_i} = Y_i - X_i \hat{\theta_i}
\end{equation}. For our application, set $Y$ is our observed data and set $X$ is the cost parameter variable to be estimated. 

Also, \cite{wang2009fast} has suggested a RLS algorithm based on both constant and variable forgetting factor.
After the least square method, the estimate obtained at time $t$ is
\begin{equation}\label{eq:Lsmw5}
  \ \hat{\theta_t} = \left( X^T_tX_t \right)^{-1}X^T_tY_t
\end{equation}
where,
$Y^T_t$ = ( $y_1$, $y_2$, ........., $y_t$),
$X^T_t$ = ( $x_1$, $x_2$, ........., $x_t$),
the estimation of time $t+1$ is calculated as
\begin{equation}
\left.
\begin{aligned}
    \hat{\theta_{t+1}} &= \hat{\theta_t} +        K_{t+1}\left(y_{t+1} -x^{T}_{t+1}\hat{\theta_t}\right) \quad\\
     P_{k+1} &= \frac{P_t}{\lambda+x^T_{t+1}P_tx_{t+1}}\\
    K_{t+1} &= P_{t+1}x_{t+1}
\end{aligned}
\right\}
\end{equation}
where,
$\lambda$ is the forgetting factor which needs to be carefully set which is a design issue. According to \cite{wang2009fast}, for time-varying $\lambda$ a good approach is to set it to a function of estimation error $\hat{e_t}$ as follows:
\begin{equation}\label{eq:lambda_value}
\lambda = 1-\alpha_1\left(\frac{1}{\pi} \arctan\left(\alpha_2\left(|\hat{e_t}|-\alpha_3\right)\right)+\frac{1}{2}\right)
\end{equation}
where, 
$\alpha_1$, $\alpha_2$ and $\alpha_3$ are all design parameters. 
\begin{multline}
\label{eq:lambda_vary}
  \lambda_t=\begin{cases}
               1-\frac{\alpha_3}{\pi}arctan(\left|{R_t-1}\right|), if \left|{R_t-1}\right| \geq \alpha_2;\\
               \alpha_1+\frac{1}{\pi}(1-\alpha_1)(arctan(1-\left|{R_t-1}\right|)) else;
            \end{cases}\\
            R_t = \begin{cases}
            max(\frac{\theta^{ij}_{t-k}}{\theta^{ij}_t},
            \frac{\theta^{ij}_t}{\theta^{ij}_{t-k}}),if \theta^{ij}_{t-k}\theta^{ij}_t \neq 0 \\
            \infty else.
            \end{cases}
\end{multline}\\
$\forall i \in (1,2,...,n)$, $\forall j \in (1,2,..,m)$, with $k,\alpha_1,\alpha_2,\alpha_3$ tunable parameters, $\frac{1}{3}\leq \alpha_1 < 1$, $\alpha_2 \geq 0$, $0\leq \alpha_3\leq 2$, $k \in N$\\

In equation~\ref{eq:lambda_vary}, the authors have suggested a time varying forgetting factor which enables more accurate tracking than constant forgetting factor given in equation~\ref{eq:lambda_value}.
Here also, for our application, set $Y$ is our observed data and set $X$ is the cost parameter variable to be estimated. 

 The KF is often used when the parameter is linearly time-varying, where the equation of the system model is given as:
\begin{equation}
\begin{aligned}
x_{k+1} &= A_kx_k+B_ku_k+Gw_k\\
y_k &= C_kx_k + v_k\\
\end{aligned}
\end{equation}
where,
$x$ is the parameter to be estimated and $y$ is the observation of $x$. Also, $x(k) \in \mathbf{R^n}$, $u(k) \in \mathbf{R^n}$, $w(k) \in \mathbf{R^n}$, $v(k) \in \mathbf{R^r}$ and $y(k) \in \mathbf{R^r}$. Moreover, $w(k)$ and $v(k)$ are white, zero mean, Gaussian noise. The KF results from the recursive application of the prediction and the filtering cycle as given by the following equations: 
\begin{equation}
\label{prediction_cycle}
\begin{aligned}
\hat{x}(k+1\mid{k}) &= A_k\hat{x}(k\mid{k}) + B_k u_k\\
\hat{P}(k + 1\mid{k}) &= A_kP(k\mid{k})A^T_k\\
\end{aligned}
\end{equation}
\begin{equation}
\label{correction_cycle}
\left.
\begin{aligned}
\hat{x}(k\mid{k}) &= \hat{x}(k\mid{k-1}) + K(k)[y(k)-C_k\hat{x}_{k\mid{k-1}}]\quad\\
K(k) &= P(k\mid{k-1})C^T_k[C_kP(k\mid{k-1})C^T_k  +R]^{-1}\\
P(k\mid{k}) &= [I -K(k)C_kP(k\mid{k-1})]\\
\end{aligned}
\right\}
\end{equation}
where, $\hat{x}(k+1)$ is the new estimation for the variable $x(k)$. Equation~\ref{prediction_cycle} gives the prediction cycle of the KF dynamics and equation~\ref{correction_cycle} gives the filtering cycle . Here, $K(k)$ is the KF gain. Here, in our work, $C(t,e,b,f,o)$ is the parameter which is to be estimated. Thus, $x$ in the general KF equation, is $C(t,e,b,f,o)$ in our application. 

We deploy LSMW method \cite{wang2009fast} first because it is a naive, inexpensive approach to estimate variables on-line. We also deploy the RLS algorithm proposed in  \cite{wang2009fast} with time-varying forgetting factor (equation~\ref{eq:lambda_vary}) as it helps in accurate tracking. Thereafter, we deploy KF method to further enhance the accuracy. 

\section{Experimentation Set-up}
\label{exp_setup}
The experimental set-up is conceptualized like a prototype transportation platform (section \ref{prototype}) where the AGVs perform specific tasks.
\subsection{Details of Set-up}
\label{detail_setup}
The platform is conceptualized consisting of three AGVs, each of them have the $L$0.0 and $L$0.1 levels implemented and all of them are individually controlled by their $L$1 level. Now, each of the AGVs have been given to traverse three different paths in the environment (Figure~\ref{netlogosample}). Here, traversing a particular path is an example of doing a task. 
For our experiment, we have assigned 2 different types of arcs to each of these three paths. Therefore, each of the robot will have 2 different $C(t,e,b,f,o)$ parameters to be estimated. For current scenario, the effect of discharge of batteries and effect of the frictional force of the floor are considered. Thus, $C(t,e,b,f,o)$ is re-formulated as $C(t,e,f)$. 

As in Figure~\ref{netlogosample}, $AGV_1$ is required  to go from $n_7$ to $n_{12}$ and there are 2 arcs in that path. So there are 2 cost parameters for $AGV_1$ namely, $C_{79}\left(t,e,f\right)$ and $C_{912}\left(t,e,f\right)$. Similar is the case for both the other AGVs ($AGV_2$ and $AGV_3$). 
\subsection{Experiment-I}
\label{exp1}
For the time being, the estimation method is applied to estimate one such cost parameter variable like $C_{79}\left(t,e,f\right)$ or $C_{912}\left(t,e,f\right)$ to validate whether the standard estimation methods, generally used for position or mechanical parameter estimation, actually works for time variable estimations of these cost parameters in order to predict the next values of these at the run-time. Let that cost parameter be denoted as $C_0\left(t,e,f\right)$. The $C_0\left(t,e,f\right)$ is measured till the battery of the mobile robot drains out and the robot comes to complete halt. The variable $C_0\left(t,e,f\right)$ is dependent only on the state of the battery charge and the frictional force of the floor where the robot is moving. 

\begin{figure}
\centering
\includegraphics[scale = 0.4]{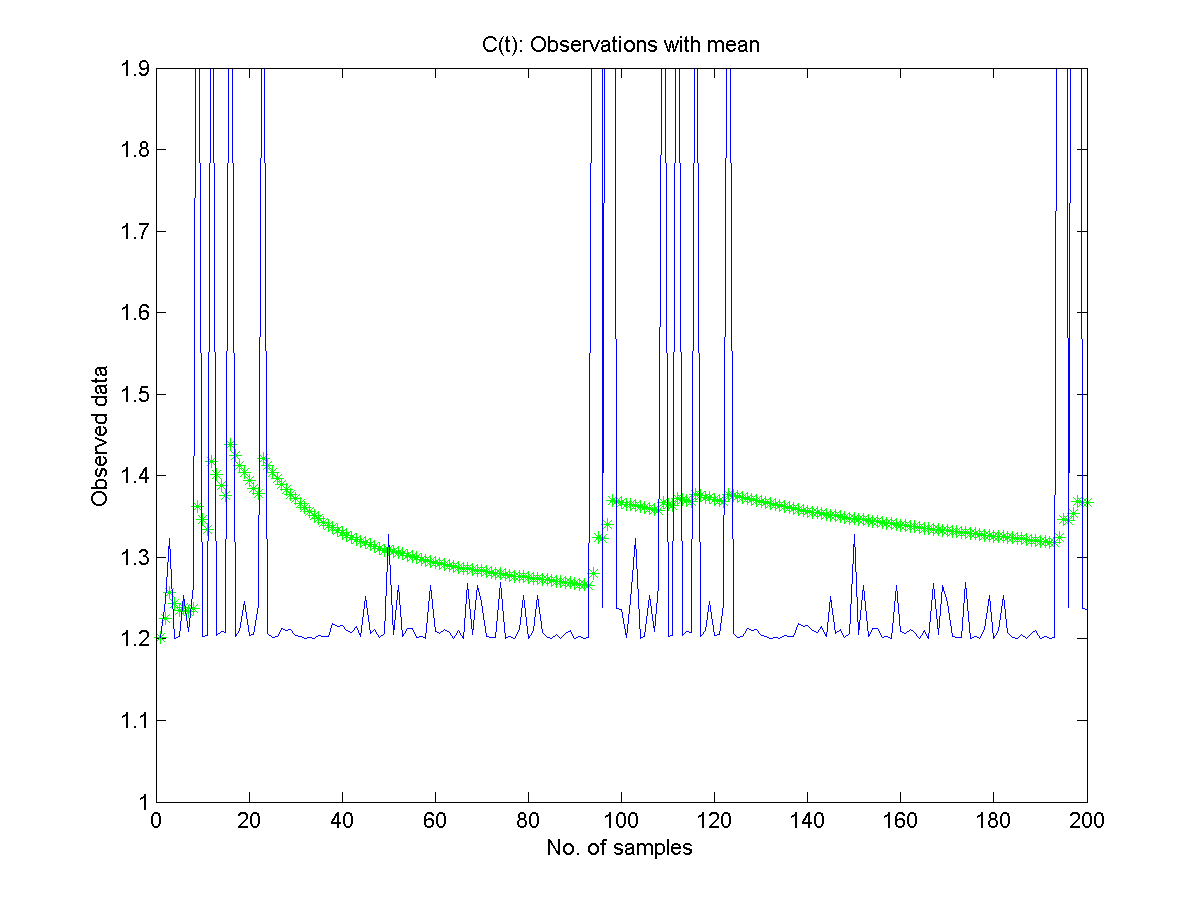}
\caption{The observed $C_0\left(t,e,f\right)$}
\label{measure_c(t)}
\end{figure}

The plot of the observed (measured) $C_0\left(t,e,f\right)$ over time is given in Figure~\ref{measure_c(t)}. Several epochs of tests are done to measure the same $C_0\left(t,e,f\right)$ in order to ensure that we are measuring the right data. 

\begin{figure}
\centering
\includegraphics[scale = 0.3]{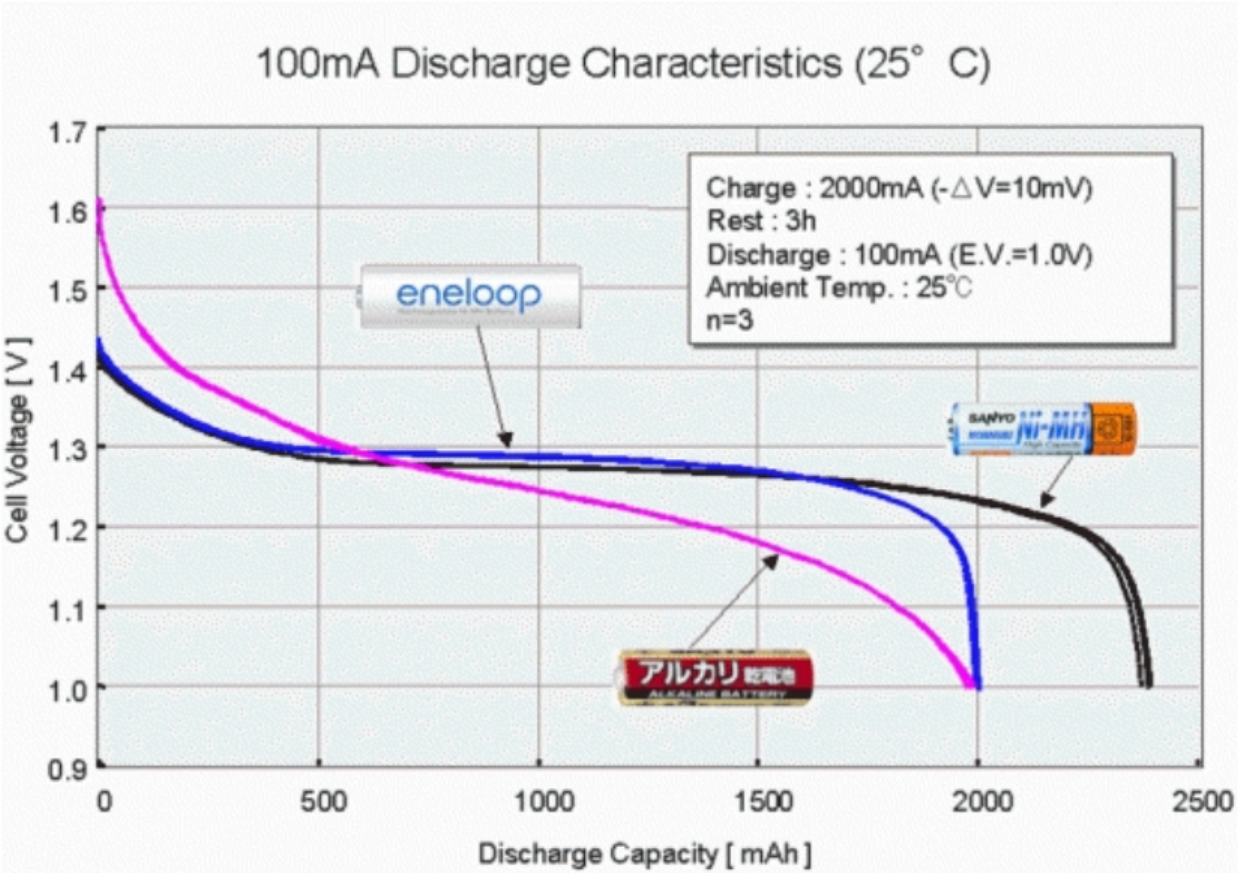}
\caption{The battery discharge profile}
\label{battery}
\end{figure}
This plot shows that $C_0\left(t,e,f\right)$ is influenced by the battery (\textit{ENELOOP} batteries for operation) discharge capacity, which is given in Figure~\ref{battery}, where the cell voltage is plotted corresponding discharge capacity. In Figure~\ref{battery}, we can observe that the cell voltage first starts from a high value and falls steeply to a steady value for a long time till the end of getting completely discharged. Now when we compare our observed data for the cost parameter. We can observe that the values start from the high magnitude and follow a steady fall during the initial stage then follows a constant value for a span of time and then again rises up to a big magnitude before the robot finally stops moving. So  $C_0\left(t,e,f\right)$ also has a similar nature as that of the battery cell voltage, but not exact nature. So we can say, in addition to the effect of battery power, the cost parameter ($C_0\left(t,e,f\right)$) is influenced by the roughness or the smoothness of the floor.

\begin{figure}
\centering
\includegraphics[scale = 0.4]{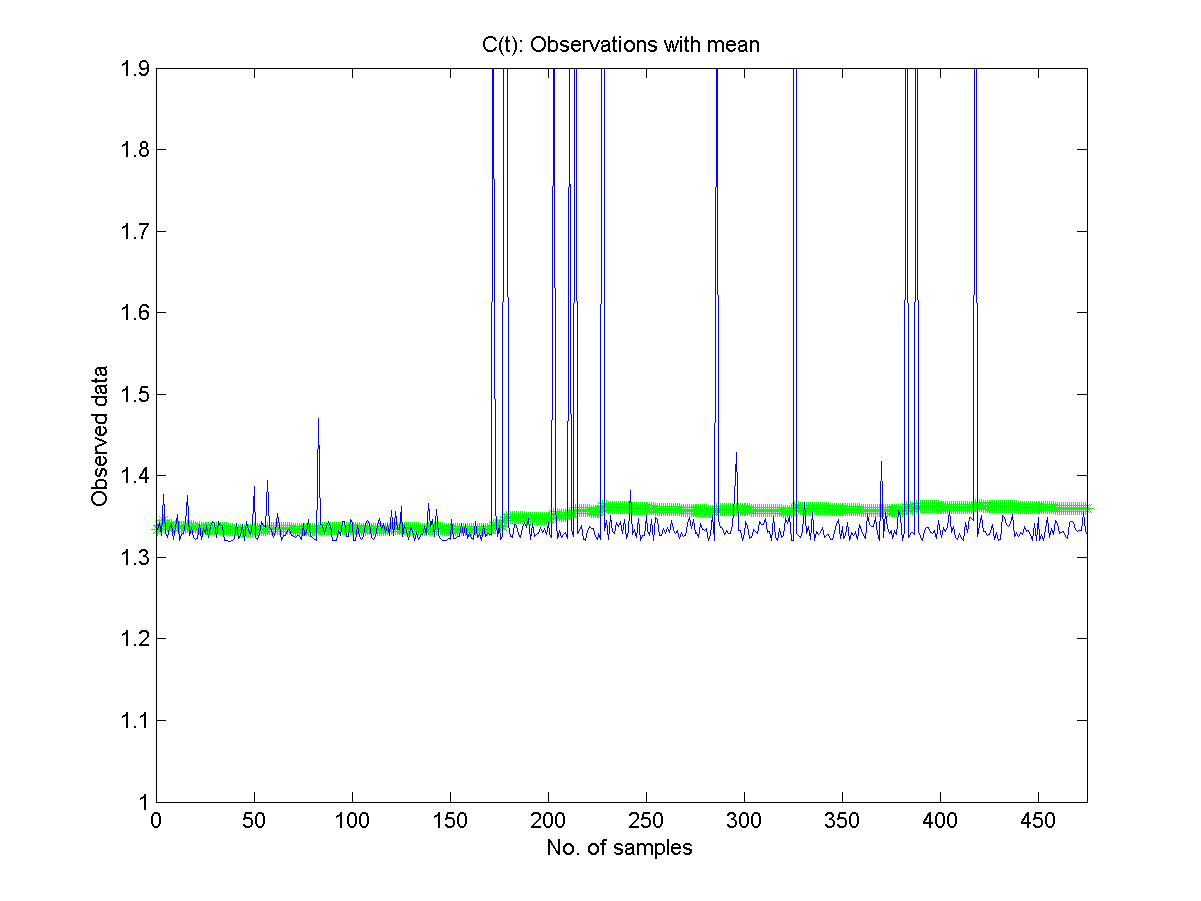}
\caption{The observed $C_{79}\left(t,e,f\right)$}
\label{second_measure_c(t)}
\end{figure}

To show that, we consider another observed data of another cost parameter in a different condition of shop floor, for example $C_{79}\left(t,e,f\right)$ (detail of this is explained in section \ref{detail_setup}). In Figure~\ref{second_measure_c(t)}, when the same batteries are used, although the nature of the observed data is similar to battery power, but the differences of observed data between $C_0\left(t,e,f\right)$ and $C_{79}\left(t,e,f\right)$ arise from the fact that the condition of floors are different.  

Hence, we can say there are cost parameters involved in performing ability of the mobile robots. Thereafter, we applied the different methods of parameter estimation to estimate one of these cost parameters ($C_0\left(t,e,f\right)$).


\subsection{Experiment-II}
\label{exp2}
\begin{figure}
\centering
\includegraphics[scale = 0.135]{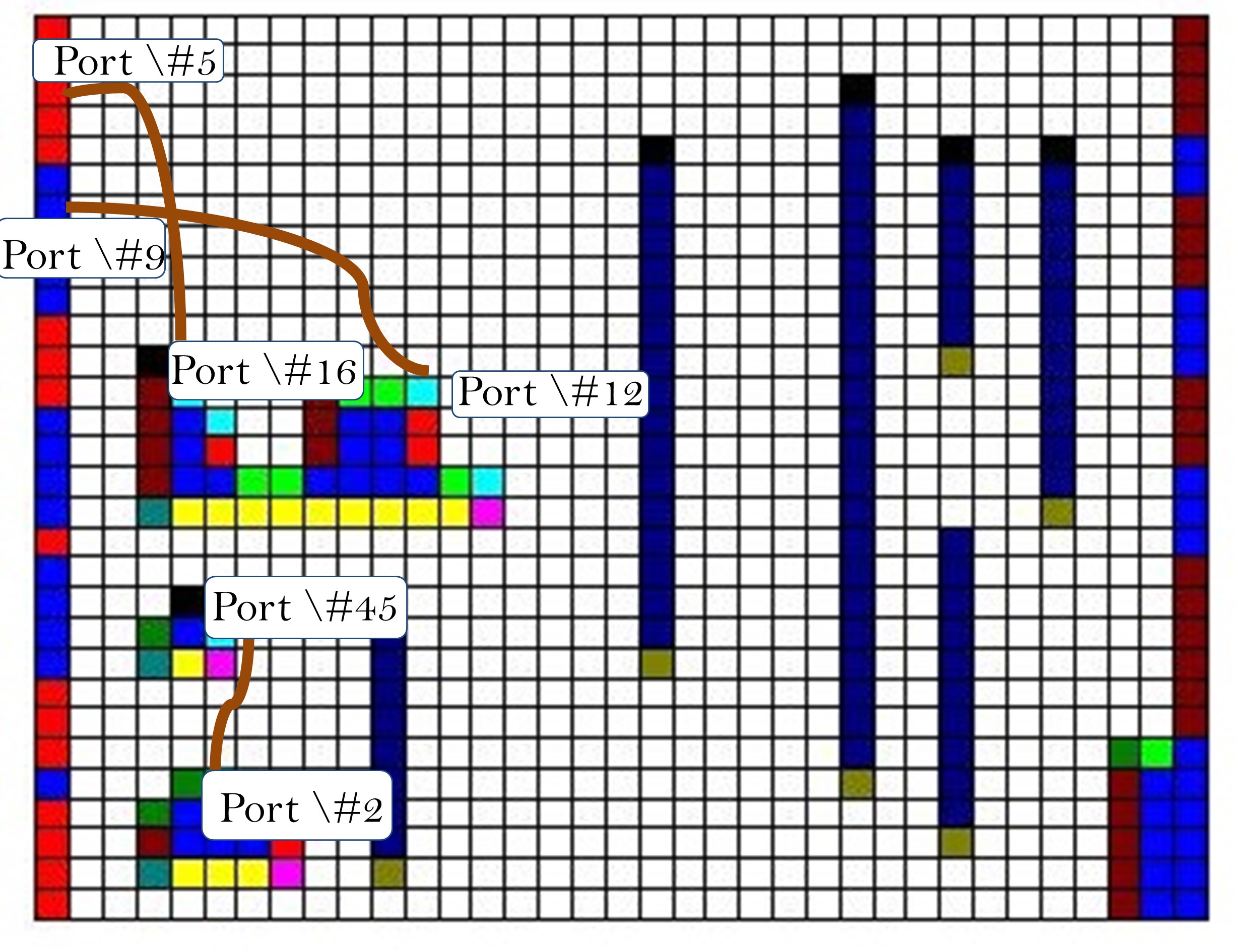}
\caption{The map of the floor}
\label{map}
\end{figure}

In our next experiment, we created a shop floor made up of the topological map given in Figure \ref{map}. The colored boxes shows the ports and the blank boxes are free-ways. The surface of the floor were made different in several places to induce variability in the cost parameters. There are several arcs marked from different ports. For example, the $Arc_{516}$ connects the Port\#5 and Port\#16, $Arc_{912}$ connects the Port\#9 and Port\#12. In this experiment, we consider the traversal time of these arcs as realistic costs as because the traversal time is previously parametrised as a cost parameter in this work. We estimate different cost parameters of each arc and compare with fixed heuristics cost like Euclidean distance from one port to another. In Table \ref{tab:PPer} in Section \ref{res2}, the comparison is described and the analysis is given in Section \ref{ananlysis2}.

\section{Results of the Experiments}
\subsection{Results of deploying estimation methods}
\label{res1}
\begin{figure}
\centering
\includegraphics[scale = 0.5]{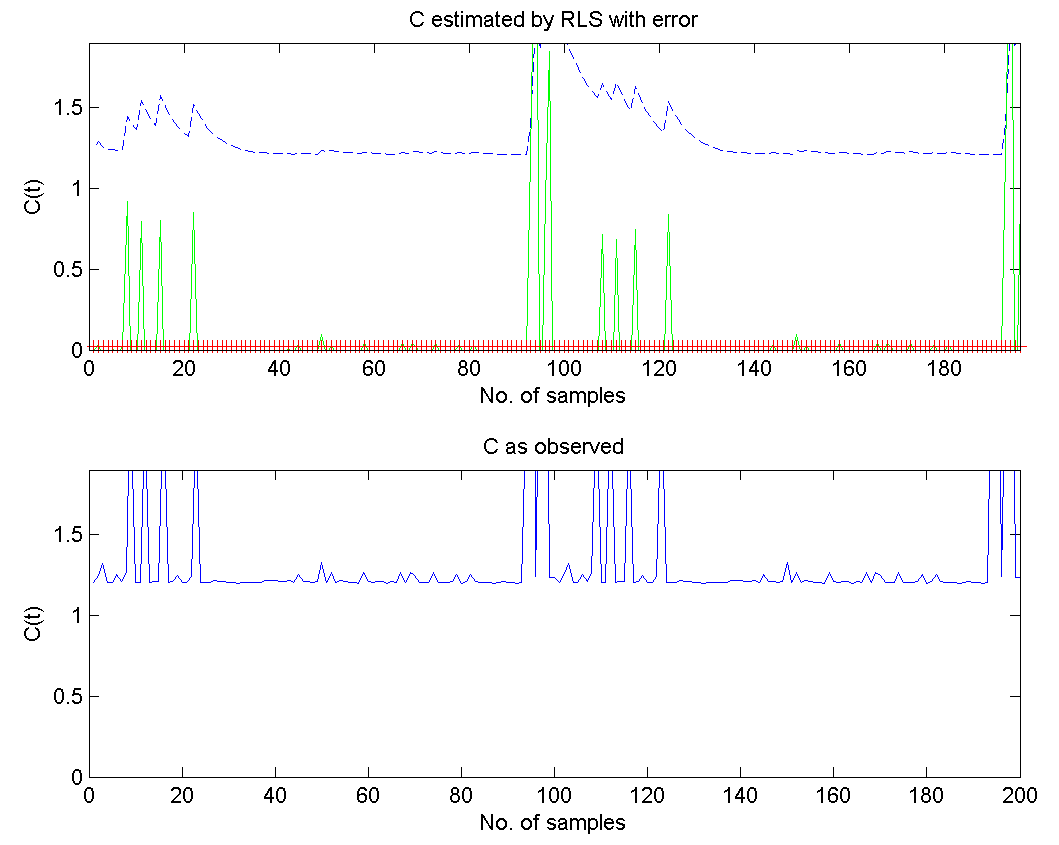}
\caption{Recursive Least Square estimation}
\label{RLS}
\end{figure}

\begin{figure}
\centering
\includegraphics[scale = 0.5]{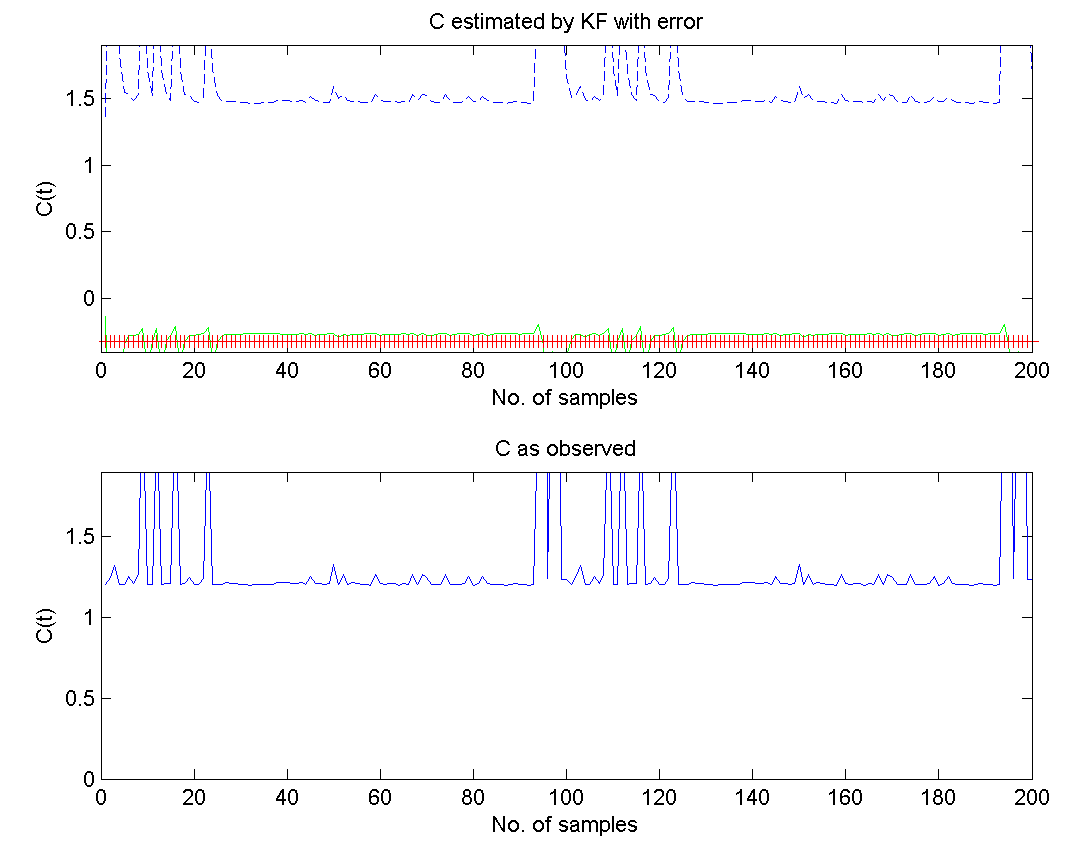}
\caption{Kalman Filtering estimation}
\label{KF}
\end{figure}

As discussed in Section \ref{exp1}, we applied three parameter estimation methods to estimate one of the cost parameters. In all the plots of results of these methods, the upper part shows the estimation result of the method and the lower part shows the observation. Also, in all the plots, blue line shows the estimated values, the green line shows the error values and the red thick line shows the mean error value. At first the Least square moving window (LSMW) method is implemented to find the estimation. The window size was taken as 5.
\begin{figure}
\centering
\includegraphics[scale = 0.5]{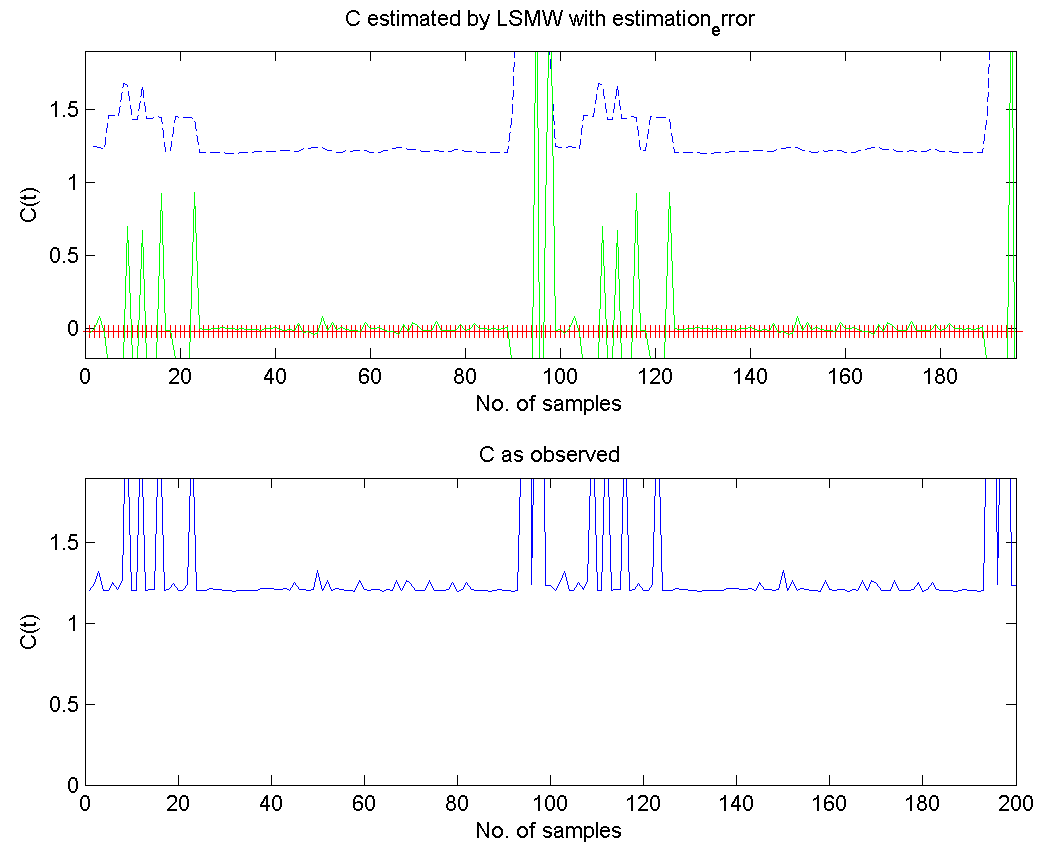}
\caption{Least Square Moving Window estimation}
\label{LSMW}
\end{figure}
In Figure~\ref{LSMW}, the results of estimates obtained by LSMW method demonstrated at the top and in the bottom the observation is plotted. In the upper part of the plot in Figure~\ref{LSMW}, the mean of the estimates are observed to be of the order of $10^(-1)$. The error level is being reduced by applying the Recursive Least Square (RLS) method, whose estimation results are shown in Figure~\ref{RLS}. As we can see, the mean of the estimation error by RLS method is further reduced to $10^(-2)$. The results of estimation done by the Kalman filtering is given by the Figure~\ref{KF} and the mean of the estimation error is of the order of $10^(-3)$. All the computations done are enough short in time to be obtained in real time.


\subsection{Discussion of the results for Experiment-I }
\label{ananlysis1}

\begin{figure}
\centering
\includegraphics[scale = 0.5]{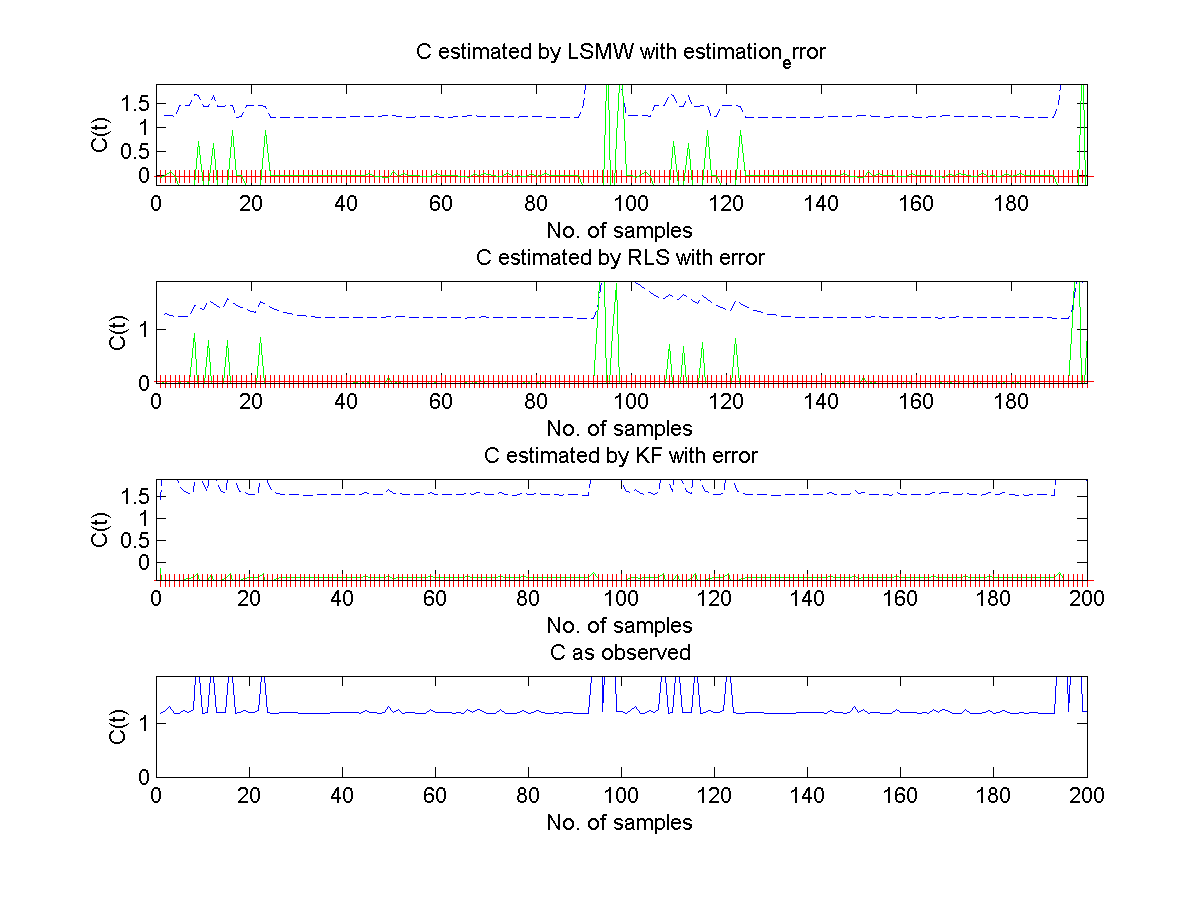}
\caption{Recursive Least Square estimation}
\label{all}
\end{figure}

 
Conclusively, we can observe in Figure~\ref{all}, that in LSMW method the first lap of values are well estimated, but the estimates fall flat to the zone of less variability. Also, the sharp variations are not estimated well by LSMW method. RLS method is able to mitigate the shortcomings of LSMW method where the sharp rise zone in the middle is estimated well. Also, it is evident Figure~\ref{all}, that Kalman Filtering provides the best estimates for estimating the time-varying cost parameter where both the sharp rise and less variable zones are estimated well. Henceforth, we can infer that the Kalman Filtering method provides the most suitable estimation for the cost parameter. These predictions are obtained on-line during the run-time of the multi-robot prototype system.


  

\subsection{Results of Experiment-II}
\label{res2}

In general, heuristics cost are mostly used in the co-operative and collaborative decision making like navigation, task allocation, obtaining optimal path. In Table~\ref{tab:PPer}, the comparison between these heuristics cost and the real estimated cost are provided. The table shows first the heuristics cost of traversing three arcs $Arc_{516}$, $Arc_{245}$  and $Arc_{912}$ ( arcs are explained in Section \ref{exp2}) and second the real cost of traversing and third how the real estimated cost varies over a single factor like battery power. A detail discussion of this comparison is provided in Section \ref{ananlysis2}. 
\begin{table}[ht]
\caption{Analysis of real costs}
\begin{adjustbox}{width=0.5\textwidth}
\centering 
\begin{tabular}{l l r c rrrrr} 
\hline\hline 
 No &\multicolumn{2}{c}{Euclidean Cost} & 100\% Battery Level &\multicolumn{5}{c}{Costs per Battery Level}
\\ 
& & & Realistic Static Cost &\\ 
\hline
& & &  & 90\% & 80\% & 70\% & 60\% & 50\% \\
\hline
&[cm] & [s] & [s] & [s] & [s] & [s] & [s] & [s]\\
\hline
$Arc_{516}$ & 30 & 1.5 & 1.7 & 1.7 & 1.9 & 2.3 & 2.5 & 2.8  \\[1ex]
\hline
$Arc_{245}$ & 17 & 1.2 & 1.3 & 1.5 & 1.5 & 1.8 & 1.97 & 2 \\[1ex]
\hline
$Arc_{912}$ & 45 & 2 & 2.2 & 2.2 & 2.4 & 2.7 & 2.9 & 3  \\[1ex]

\hline\hline 
\end{tabular}
\label{tab:PPer}
\end{adjustbox}
\end{table}

\subsection{Discussion of the results for Experiment-II }
\label{ananlysis2}
In the Table~\ref{tab:PPer}, it is observed that estimated values of real cost is varying over the battery discharge level. But, the time-varying nature of costs over different factors is generally not included in heuristics costs. In Table~\ref{tab1:Per} the percentage differences between the heuristics cost and the cost parameter computed from one-time characterization in reality and progressive computation is demonstrated. 
So, from the onset of functioning of the multi-robot system, the costs of performing efficiently varies over the time and correct estimates of them needs to be utilised as it is evident from the comparison. Therefore, these real estimated costs can be utilised to make more cost efficient co-operative decisions in the system controller for each AGV. 
\begin{table}[ht]
\caption{Difference between Computed Real Costs and Heuristics Costs}
\begin{adjustbox}{width=0.5\textwidth}
\centering 
\begin{tabular}{l l rrrrr} 
\hline\hline 
 No & Between Heuristics and on-time real cost &\multicolumn{5}{c}{Difference of Real and Heuristics Costs per Battery Level}
\\ 
\hline
& & 90\% & 80\% & 70\% & 60\% & 50\% \\
$Arc_{516}$ & 13\% & 13\% & 26\% & 53\% & 66\% & 86\%  \\[1ex]
\hline
$Arc_{245}$ & 8.3\% & 25\% & 25\% & 55\% & 64\% & 66.4\% \\[1ex]
\hline
$Arc_{912}$ & 10\% & 10\% & 20\% & 35\% & 40\% & 50\%  \\[1ex]

\hline\hline 
\end{tabular}
\label{tab1:Per}
\end{adjustbox}
\end{table}

 In Section \ref{background} we have presented two works which approach a nearly similar problem \cite{nestinger2012adaptive, Confessore2011}, but their work cannot be directly compared to ours because the approaches are different (also described in Section \ref{background}). Moreover, to best of our knowledge, there is no other state of the art research proposal which addresses the same problem as we are solving. So we cannot present a comparison of our work with state of the art.

\section{Conclusion}
The concept of cost parameters in the multi-robot system implemented in transportation and automation industry is formulated and efficient estimation method of these is proposed. Moreover, the necessity of these cost parameter estimation is shown which can help to generate better control decisions. It is evident from the results that Kalman filtering provides the best method for estimating the proposed cost parameters. In the topological map for the shop floor, proposed cost parameters are timed linked to each arc of the map. Also, the results show that the heuristics cost differ from the cost parameter in reality when computed one-time by 9\% and when computed progressively by maximum 67\%. 

In our current work, a single task is considered for each AGV. Here, 'single task' means repeating the same path over and over again. However, in real industrial scenario, each AGV is capable of performing multiple tasks for improving the system performance. Hence, in the progress of our work, the cost parameters are to be considered in multiple task performing scenario for each AGV when all the assigned tasks like carrying a load, traversing a path, fitting a mechanical part etc for one AGV will be interleaved with one another. The nature of these cost parameters will change for each AGV. Based on the estimates of the cost parameters, more optimal and cost efficient task allocation decision can be taken at the system level controller. Therefore, this work paves the way for future cost estimations with interleaving tasks for each mobile robot and the group of mobile robots for completing one or more tasks. Moreover, the real time values of these cost parameters are to be used and updated in the knowledge base of the system used for characterization. This implies the data obtained from the mobile robot about the parameters can be used to predict the future time and energy to do perform similarly.






%

\AtEndEnvironment{thebibliography}{
\bibitem{ribas2013agent}
 Removed for blind review
 
\bibitem{LluisIsmaelMoreno}Removed for blind review

}

\bibliographystyle{plain}
\bibliography{DR_ETFA16.bib}
\end{document}